\documentclass[10pt, conference, compsocconf]{IEEEtran}
\IEEEoverridecommandlockouts
\usepackage{multirow}
\usepackage{color}
\usepackage{subfigure}
\usepackage{graphicx}
\usepackage{epsfig}
\usepackage{subfig}
\usepackage{algorithm}
\usepackage{algorithmicx}
\usepackage{algpseudocode}
\usepackage{epstopdf}
\usepackage{amsmath}
\usepackage{url}

\begin{document}
%
\title{Multiple Object Tracking with Kernelized Correlation Filters in Urban Mixed Traffic}

\author{\IEEEauthorblockN{Yuebin Yang*}
\IEEEauthorblockA{University of Electronic Science and Technology of China\\
Chengdu, China\\
Email: patrick.ybyang@gmail.com}
\and
\IEEEauthorblockN{Guillaume-Alexandre Bilodeau}
\IEEEauthorblockA{LITIV lab. Dept. of computer \& software eng.
\\Polytechnique Montr\'eal, Canada\\
Email: gabilodeau@polymtl.ca}
\thanks{*This work was conducted while Yuebin Yang was doing a MITACS Globalink internship at Polytechnique Montr\'eal}
}

\maketitle

\begin{abstract}
Recently, the Kernelized Correlation Filters tracker (KCF) achieved competitive performance and robustness in visual object tracking. On the other hand, visual trackers are not typically used in multiple object tracking. In this paper, we investigate how a robust visual tracker like KCF can improve multiple object tracking. Since KCF is a fast tracker, many KCF can be used in parallel and still result in fast tracking. We built a multiple object tracking system based on KCF and background subtraction. Background subtraction is applied to extract moving objects and get their scale and size in combination with KCF outputs, while KCF is used for data association and to handle fragmentation and occlusion problems. As a result, KCF and background subtraction help each other to take tracking decision at every frame. Sometimes KCF outputs are the most trustworthy (e.g. during occlusion), while in some other cases, it is the background subtraction outputs. To validate the effectiveness of our system, the algorithm was tested on four urban traffic videos from a standard dataset. Results show that our method is competitive with state-of-the-art trackers even if we use a much simpler data association step.
\end{abstract}


\IEEEpeerreviewmaketitle

\section{Introduction}
Multiple object tracking (MOT) is a fundamental task in computer vision with numerous conceptually diverse tracking algorithms being proposed every year. A popular application area is traffic surveillance because of its practical use for reducing traffic jam and for assessing the security of various road configurations. However, the existing intelligent transportation systems (ITS) exhibit a low performance when faced with problems like occlusions, illumination changes, motion blur and other environmental variations \cite{Lessard_2016_CVPR_Workshops}. To address these problems, we propose a multiple object tracker based on a robust visual object tracker. Using a visual object tracker allows our method to handle occlusions in a straight-through manner (objects are tracked individually during occlusions) even if the detected objects are obtained as blobs from background subtraction. This contrast with the typical merge-split approach used by recent trackers based on background subtraction \cite{JodoinITS2016,Torabi2009}. Furthermore, the use of a robust visual tracker reduces the need for a complex data association method. Our proposed method is online meaning that it takes decision based on past frames only.

In this paper, we combine background subtraction with KCF (Kernelized Correlation Filters) tracker \cite{HenriquesKCF} and propose a promising model-free MOT method for road users in urban mixed traffic. Our method is based on background subtraction because urban mixed traffic can include unexpected objects. As such, a trained object detector is difficult to design so that it does not miss objects. Background subtraction provides us with noisy candidate object regions. We analyze and process each candidate region to get higher-quality potential objects. For tracking, we create a KCF tracker for each candidate region using a grayscale and color names appearance model. Sequentially, moving objects are tracked and their states are updated frame by frame. To handle scale variations, object fragmentation, occlusions and lost tracks, we use simple data association between KCF trackers and targets detected by background subtraction. Object states are determined by using the information of both KCF and background subtraction as they can both make errors at different times. Finally, the system is evaluated with four different urban videos involving cars, pedestrians, bikes, and trucks. Results show that our method is competitive even if using simple data association, demonstrating the benefit of using a strong visual object tracker in a MOT framework.

This paper is organized as follows: In section \ref{rwork}, we discuss related work, in section \ref{metho}, we present our proposed method, including extracting regions of interest (ROI), background subtraction, foreground analysis, and the tracking of objects. In section \ref{experiments}, we test our method and summarize our results. Finally, in section \ref{conclusion}, we conclude the paper. 
     
\section{Related work}
\label{rwork}
For tracking road users, there exist basically two approaches depending on the way the road users are detected, which is either by using optical flow, or by using background subtraction. Methods based on pre-trained bounding box detectors are not applied in road user tracking scenarios because it is difficult to design a universal detector that can detect all possible road users from various viewpoints. Objects may be missed. The use of background subtraction and optical flow is not without limitation either, because although all objects are usually detected, they are often fragmented into two or more parts, or they can be merged.   

\subsection{Optical flow-based methods}
In this family of methods, objects are detected using optical flow by studying the motion of tracked points in a video. Feature points that are moving together are considered as belonging to the same object. Several methods accomplished this process using the Kanade-Lucas-Tomasi (KLT) tracker \cite{KLT}. Among others, the trackers of Beymer \textit{et al.}~\cite{Beymer1997}, Coifman \textit{et al.}~\cite{coifman_real-time_1998}, Saunier \textit{et al.}~\cite{saunier_feature_based_2006} and Aslani and Mahdavi-Nasab~\cite{Aslani2013}. For example, Saunier \textit{et al.}~\cite{saunier_feature_based_2006} method, named \textit{Traffic Intelligence}, tracks all types of road users at urban intersections by continuously detecting new features and adding them to current feature groups. Then, the right parameters should be selected to segment objects moving at similar speeds, while at the same time not over-segmenting smaller non-rigid objects such as pedestrians. Because objects are identified only by the motion of feature points, nearby road users moving at the same speed may be merged together. Furthermore, the exact area occupied in the frame by the road user is unknown because it depends on the position of sparse feature points. Finally, when an object stops, its features flow interrupts which usually leads to fragmented trajectories.     

\subsection{Background subtraction-based methods}
This second family of methods rely on background subtraction to detect road users in a video. Background subtraction gives blobs that can correspond to parts of objects, one object, or many objects grouped together. The task is then to distinguish between merging, fragmentation, and splitting of objects. This approach works fairly well under conditions with no or little occlusion. However, under congested traffic conditions or slow speed conditions, vehicles are harder to extract from the background as they may partially occlude with each other and many objects may merge into a single blob. In this case, it is hard to divide them.  

Examples of trackers based on background subtraction include the work of Fuentes and Velastin~\cite{Fuentes2006}, Torabi~\textit{et al.}~\cite{Torabi2009}, Jun \textit{et al.}~\cite{jun_tracking_2008}, Kim \textit{et al.}~\cite{kim_real_2008}, Mendes \textit{et al.}~\cite{mendes15vehicle}, and Jodoin \textit{et al.}~\cite{JodoinITS2016}.   

Fuentes and Velastin~\cite{Fuentes2006} proposed a method that performs simple data association via the overlap of foreground blobs between two frames. In addition to matching blobs based on overlap, Torabi~\textit{et al.}~\cite{Torabi2009} validated the matches by comparing the histograms of the blobs and by verifying the data association over short time windows using a graph-based approach. These approaches track objects in a merge-split manner as objects are tracked as groups during occlusion. 

Jodoin \textit{et al.}~\cite{JodoinITS2016,JodoinWACV14} proposed \textit{Urban tracker}, which is a tracker that in addition to background subtraction uses feature points to model the appearance of the tracked objects. This helps in solving the data association problem in the case of occlusion. The position of the bounding box of occluded objects can even be predicted based on the visible feature points and their previous positions on the object. A finite state machine is used for handling fragmentation and occlusion. The method follows initially the split-merge paradigm but after solving occlusion, backtracks and segments the occluded objects. 

Jun \textit{et al.}~\cite{jun_tracking_2008} used background subtraction to estimate the object properties. A watershed segmentation technique is used to over-segment the vehicles. The over-segmented patches are then merged using the common motion information of tracked feature points. This allows to segment vehicles correctly even in the presence of partial occlusion. Kim \textit{et al.}~\cite{kim_real_2008} combines background subtraction and feature tracking approach with a multi-level clustering algorithm based on the Expectation-Maximization (EM) algorithm to handle the various object sizes in the scene. The resulting algorithm tracks various road users such as pedestrians, vehicles and cyclists online and the results can then be manually corrected in a graphical interface. Mendes \textit{et al.}~\cite{mendes15vehicle} proposed a method that also combines KLT and background subtraction. This approach increases its accuracy by setting \textit{in} and \textit{out} regions to reconfirm object trajectories and by segmenting regions when occlusions are detected. 

In our work, we combine KCF and background subtraction and propose a model-free MOT method that can track any types of road users without prior knowledge. The advantage of our method is that it addresses the occlusion problem by tracking object individually inside blobs of merged objects. Backtracking is not necessary, neither is the use of an explicit region-based segmentation. 

\section{Tracking Methodology}
\label{metho}
\begin{figure*}
\centering
\includegraphics[width=15.0cm]{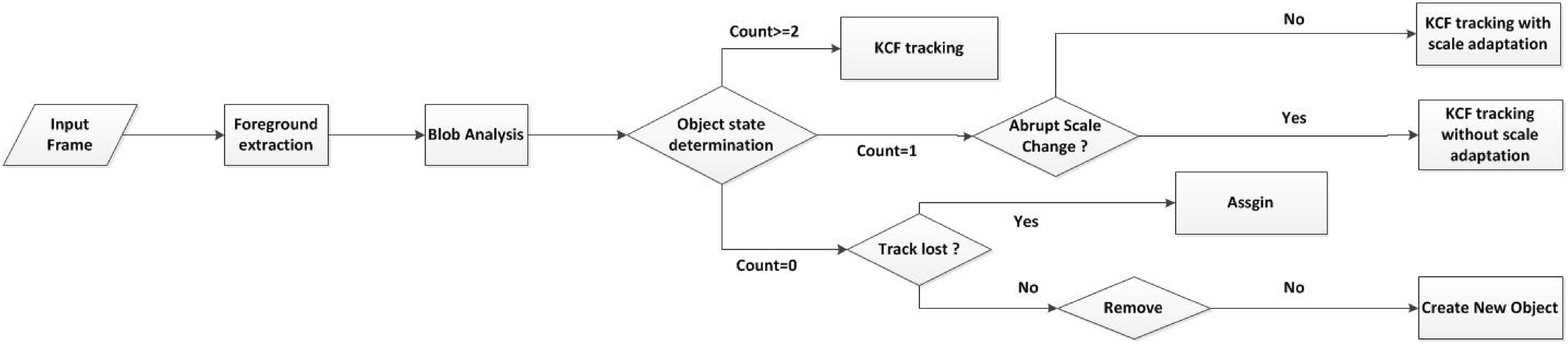}
\caption{Diagram of the steps of our proposed method} \label{fig2}
\end{figure*}

Our algorithm combines background subtraction with the KCF tracker. Background subtraction is used for extracting the foreground (candidate road users to track), and the KCF tracker is applied for data association and for tracking objects during occlusions. To improve the overall performance of our method, we added several refinements to solve problems like occlusion and scale adaptation. Figure \ref{fig2} shows the diagram of the steps of our method.  

For each frame, we extract the foreground to obtain candidate object regions (blobs). Then, we analyze the blobs and apply several filters to obtain a final list of candidate object regions $COR_i$. Next, to update the tracked object states, we compare the $COR_i$ with current tracks $CT_j$ for data association. This is done by updating all the currently active KCF trackers. The KCF trackers $KCF^{t-1}_j$ find the most plausible data association with a $COR_i$ based on proximity and their internal model. Correspondences may be many to one, one to many, one to one, or an object may be entering and leaving the scene. These different cases are determined by counting the number of KCF trackers (one per $CT_j$) associated to each $COR_i$. For objects in occlusion, we update their positions without adapting scale (KCF is more trustworthy, we use the object state provided by KCF tracker). For one to one association, we update the state of the tracked object using the information from background subtraction (Background subtraction allows adjusting the scale of the KCF tracker, and in one to one association, background subtraction is assumed to be often trustworthy). For a new object, we assign a new KCF tracker. Finally, undesired objects will be deleted if not detected by background subtraction during several frames. In the next subsections, we describe our method in detail (refer to Figure \ref{fig2}).

\subsection{Foreground detection}
Foreground detection allows us to obtain candidate object regions $COR_i$. Blobs $B_i$ are extracted at each frame. Foreground detection is used to begin new tracks and to validate the existing ones. If a detected blob $B_i$ is not associated to a KCF trackers, a new tracker will be initialized on that blob. If a KCF tracker is tracking an object and that object is not detected for many frames, this may indicate a failure of the KCF tracker. In this case, tracks are discarded. Any foreground detection method can be used with our proposed method. Since we are using a publicly available dataset with foregrounds detected using LOBSTER \cite{CharlesB14}, our method has been tested using this method.

\subsection{Blob Analysis}
\label{banalysis}
At this step, we process extracted blobs $B_i$ with several operations to get improved candidate object regions $COR_i$. Morphological operations including median filtering, closing and hole filling are applied to each $B_i$. To minimize the influence of environmental disturbance, regions smaller than a threshold $T_r$ or with inappropriate width to height ratio (e.g. objects that are too thin) are discarded. To partially solve fragmentation problems that may occur, regions that are very close spatially (based on a threshold $T_c$) are merged together. After these operations, we get the final candidate object regions $COR_i$. These regions are then used as input for tracking.

\subsection{Object Tracking}
The advantages of background subtraction are that: 1) it can detect any objects even the ones that were never observed before, and 2) it can provide scale information about the object to track. However, background subtraction also has limitations in the processing of traffic scenes: 1) under heavy traffic conditions, objects move slowly and may be integrated in the background model, thus some may not be detected correctly, and 2) when partial occlusion occurs, several objects will merge into one large blobs $B_i$ instead of several blobs. These cases can be handled more easily by the KCF tracker. Thus, our method will combine both approaches to handle these various conditions. 

KCF \cite{HenriquesKCF} is a correlation-based tracker that analyzes frames in the Fourier domain for faster processing. It does not adapt to scale changes but updates its appearance model at every frame. KCF can be used with various appearance models. In this work, we use both grayscale intensities and color naming as suggested by the authors of \cite{ExtendedKCF}.   

\subsubsection{Finding the object states} 
\label{ostatus}
Before deciding if at frame $t$ a $COR^{t}_i$ will be tracked mostly with the KCF tracker or with background subtraction, we need to determine in which condition it is currently observed. The $COR^{t}_i$ may be entering the scene, leaving the scene, isolated from other objects, or in occlusion. To determine the states of all $COR^{t}_i$, the active KCF trackers at frame $t-1$, $KCF^{t-1}_j$, are applied to the original RGB color frame $t$. Then the resulting tracker outputs $TO^{t}_j$ are tested for overlap with each $COR^{t}_i$ and vice versa:
\begin{itemize}
\item If a $COR^{t}_i$ overlaps with only one $TO^{t}_j$, it means that the object is isolated and currently tracked (\textit{state: tracked}).
\item If a $COR^{t}_i$ overlaps with more than one $TO^{t}_j$, it means that the object is tracked but under occlusion (\textit{state: occluded}).
\item If a $COR^{t}_i$ does not overlap with any $TO^{t}_j$, it means that it is an object not currently tracked (\textit{state: new object}).
\item If a $TO^{t}_j$ does not overlap with any $COR^{t}_i$, it means that it is an object not currently visible (\textit{state: invisible or object that has left}).
\end{itemize}

The overlap is evaluated using
\begin{equation}
Overlap(x,y)=\frac{x \cap y}{x \cup y},
\end{equation}
where $x$ and $y$ are two bounding boxes.

\subsubsection{State: Tracked}
\label{strack}
If a $COR^{t}_i$ overlaps with one and only one $TO^{t}_j$, we assume that the object is isolated and tracked correctly (but not necessarily precisely). Since the KCF tracker does not adapt to scale change, we add the $COR^{t}_i$ bounding box to the $CT_j^{t-1}$ associated to $KCF^{t}_j$ most of the time. This way the scale can be adapted progressively during tracking based on the blob $COR^{t}_i$ from background subtraction. In this case, we re-initialize $KCF^{t}_j$ with the information for $COR^{t}_i$ to continue the tracking. 

We use $TO^{t}_j$ only if $COR^{t}_i$ suddenly shrinks because of fragmentation of the tracked target. Indeed, background subtraction cannot always handle well abrupt environmental changes which cause over-segmentation. Therefore, when the KCF tracker bounding box is much larger than the background subtraction bounding box (as defined by thresholds $T_{ol}$ and $T_{oh}$), we assume that KCF is more reliable. Formally,

\begin{equation}
CT_j^t = 
   \begin{cases}
   CT_j^{t-1} \cup TO^{t}_j, & \text{if } T_{ol}<=\frac{A(TO^{t}_j)}{A(COR^{t}_i)}<=T_{oh} \\
   CT_j^{t-1} \cup COR^{t}_i, & \text{otherwise} \\
   \end{cases},
\end{equation}
where $A()$ is the area of a bounding box. The upper bound threshold $T_{oh}$ is used to allow adaptation of the scale when the object is leaving the scene. In this case, the object shrinks but it is not caused by over-segmentation, so the use of $COR^{t}_i$ should be considered. Note that when an object is fragmented, $KCF^{t}_j$ will be associated with the larger fragment. Therefore, the shrinking is conditioned by the larger fragment, and as a result, normally the shrinking is limited. If it is too large, most of the time it is explained by an object that is leaving the video frame. This is why $T_{oh}$ is used.

Finally, even if $COR^{t}_i$ is significantly larger than $TO^{t}_j$, we still used the background subtraction blob as shadows are not problematic in our application and $COR^{t}_i$ localization accuracy is better than $TO^{t}_j$ from KCF.  

\subsubsection{State: Occluded}
\label{soccluded}
if more than one $TO^{t}_j$ overlap with a $COR^{t}_i$, we assume that several previously isolated objects are now in occlusion. Background subtraction fails to distinguish objects as they are merged into one blob. In this case, we handle the occlusion problem in a straight-through manner and we use the bounding boxes outputted by the KCF trackers. That is, we add the $TO^{t}_j$ to the $CT_j^{t-1}$ related to $COR^{t}_i$ with
\begin{equation}
CT_j^t = CT_j^{t-1} \cup TO^{t}_j.
\end{equation}	
 
\begin{figure}
\centering
\subfigure[]{\includegraphics[width=4.0cm]{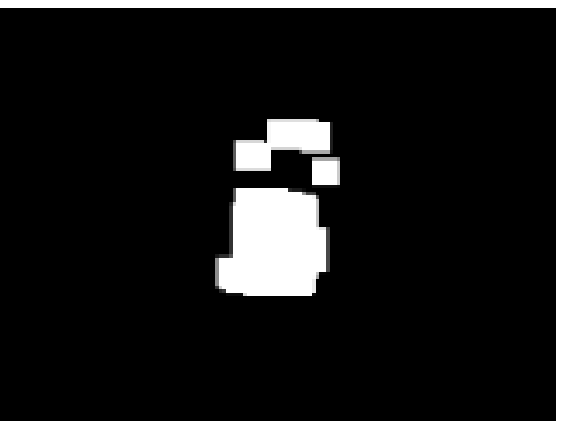}}
\label{abcccc}
\subfigure[]{\includegraphics[width=4.0cm]{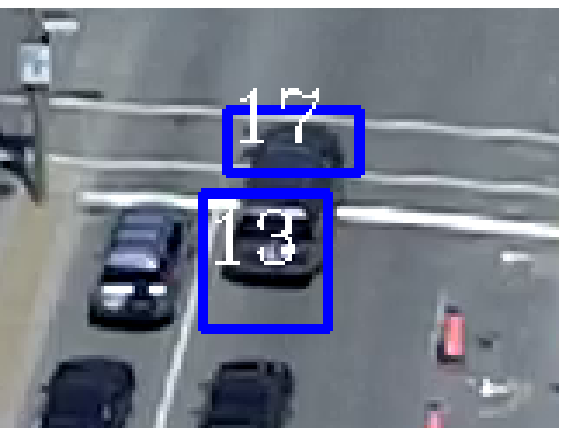}}
\subfigure[]{\includegraphics[width=4.0cm]{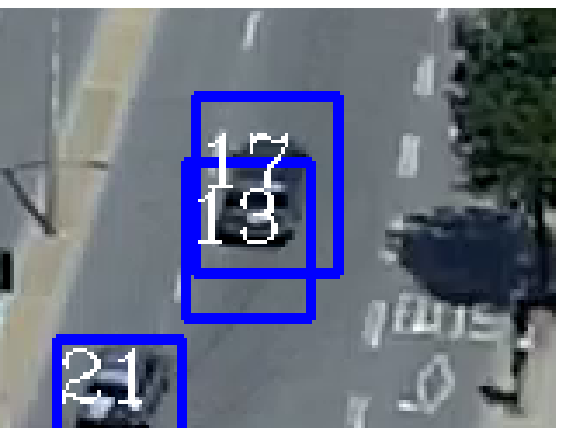}}
\subfigure[]{\includegraphics[width=4.0cm]{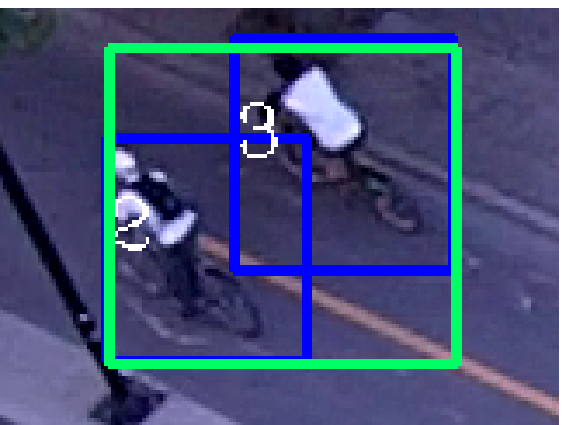}}
\caption{Example of redundant KCF trackers. (a) Extracted regions, (b) Object starts moving, (c) Redundant trackers, (d) Parallel moving objects.} 
\label{fig4}
\end{figure}

Background subtraction cannot handle over-segmentation problems in some conditions. As it can be seen from Figure \ref{fig4}(a, b, and c), when one object starts moving, redundant KCF trackers may be created on its segmentation as the object may be temporarily over-segmented. This is not a real case of occlusion. As time goes by, trackers tracking the same object may partially overlap with each other (As shown in Figure \ref{fig4}c). We proposed a method to address this problem. If the area of the blob $COR^{t}_i$ is smaller than the sum of the bounding boxes $TO^{t}_m$ and $TO^{t}_n$ of these two KCF trackers for eight consecutive frames, it is very likely to be two trackers tracking the same object. In this case, we delete the shorter-lived KCF tracker and we update the other one with $COR^{t}_i$. If not, we assume that two objects are really occluded and do not take any action (see Figure \ref{fig4}d). That is,

\begin{equation}
   \begin{array}{r l}
   \text{if }A(TO^{t}_m)+A(TO^{t}_n)>A(COR^{t}_i), & \text{Delete } KCF^{t}_n\\
   \text{otherwise}, & \text{no action}\\
   \\
   \end{array},
\end{equation}
where $A()$ corresponds to the area of a bounding box, and $KCF^{t}_n$ is the most recent of the two KCF trackers tracking the same object.

\subsubsection{State: New object}
\label{snew}
Because the KCF tracker sometimes loses track of objects when occlusion occurs, we first need to identify whether $COR^{t}_i$ is not overlapping with any $TO^{t}_j$ because of lost track or because of a possible new object. A problem that sometimes occurs is that after an occlusion two KCF trackers are on a single object, while the other object that was in occlusion is associated with no tracker. This is because of the continuous update process of KCF that can make the model drift during occlusions. In that case, a tracker should be re-assigned to the other object.  

\begin{figure}
\centering
\subfigure[]{\includegraphics[width=4.0cm]{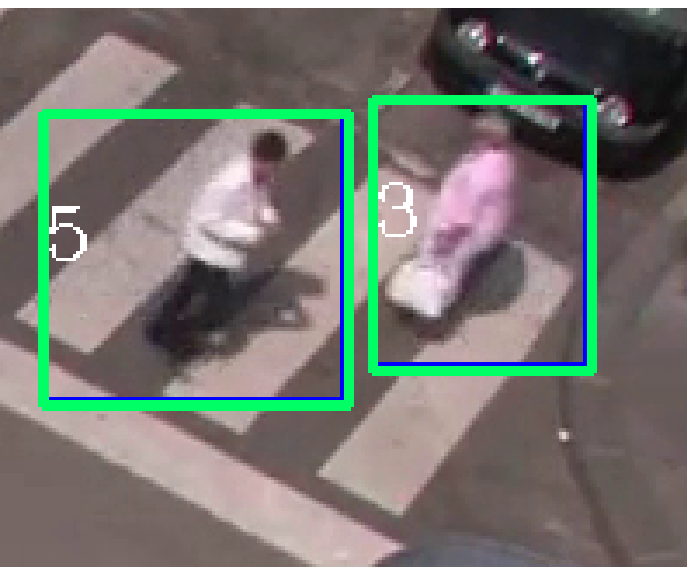}}
\subfigure[]{\includegraphics[width=4.0cm]{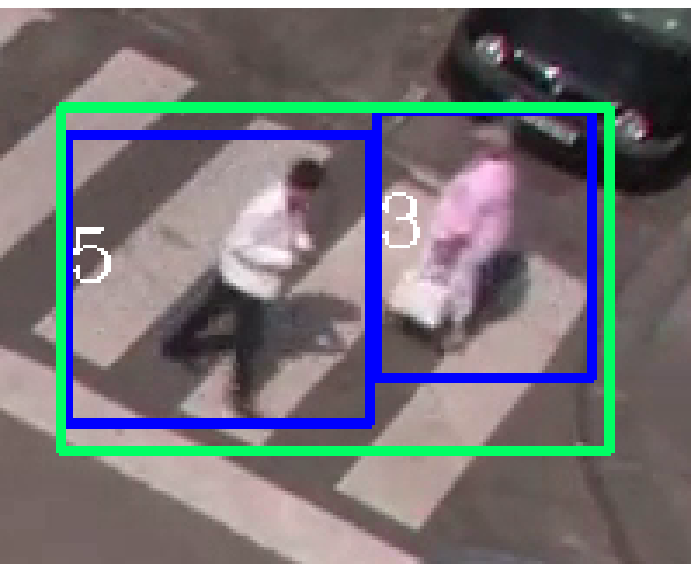}}
\subfigure[]{\includegraphics[width=4.0cm]{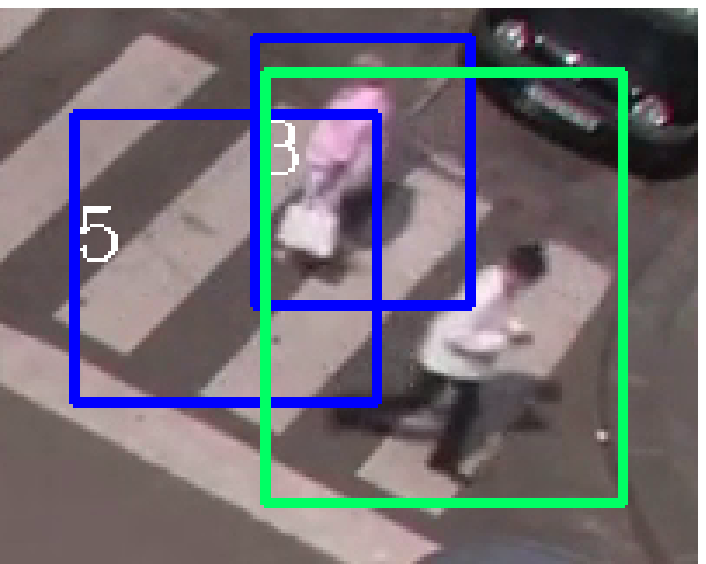}}
\subfigure[]{\includegraphics[width=4.0cm]{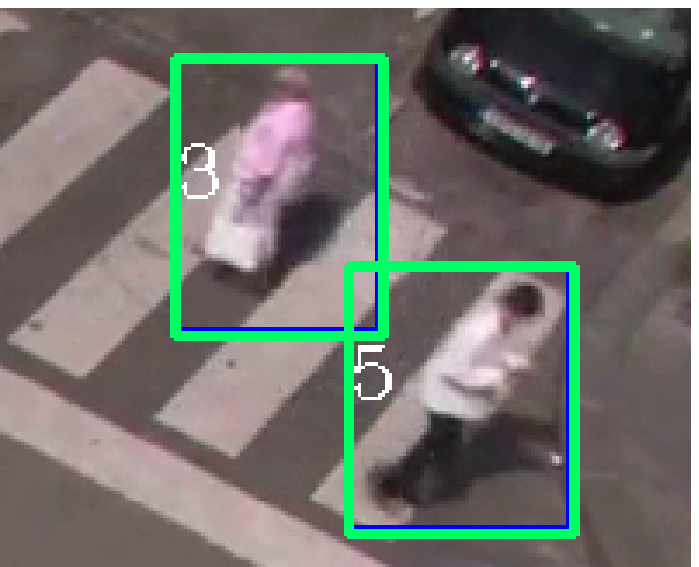}}
\caption{An example of handling tracker drift problem. (a) Before occlusion,  (b)  Group during occlusion,  (c) Two trackers on the same object, none on the other one, (d) Relabeling after split.} 
\label{fig3}
\end{figure}

To address this issue, we keep track of groups of objects. As it can be seen from Figure \ref{fig3}, when occlusion occurs, we label occluding objects as being inside a specific group. As soon as a member splits from the group without being tracked, we search for redundant KCF trackers among other group members. When more than one KCF trackers are found within a same group member, we re-assign the tracker that is matching less.

If it is indeed a new object, a new track is initialized with
\begin{equation}
CT_j^t =  COR^{t}_i.
\end{equation}	

\subsubsection{State: Invisible or exited object}
\label{sexited}
When a $TO^{t}_j$ does not overlap with any $COR^{t}_i$, we considered that the object is \textit{invisible}. If the object has exited the scene it will not reappear, but if it was hidden, it should reappear eventually. To distinguish these two cases, we use a rule based on the number of consecutive frames for which the object is \textit{invisible}. If the object is invisible for more than eight consecutive frames, it will be removed from active tracks. Based on objects average size and moving speed, eight frames equals generally to the diagonal of an object bounding box. To minimize environmental disturbances that cause false detection blobs, such as camera shake, objects that exist for less than six frames will be discarded.

To improve tracking performance, when parts of an object track are missing because the object was at time \textit{invisible}, we restore the missing parts by using interpolation from neighboring frames when the object was successfully tracked.

 Our method is summarized by algorithm \ref{algo1}.

\begin{algorithm}
  \caption{Multiple object tracking with KCF}
  \begin{algorithmic}[1]
   \State {Input: video}
   \State {Output: trajectories + bounding boxes}
    \Procedure {Tracking}{}
    \For {each video frame}

    \State {Extract foreground}	
        \For {Each extracted region} \Comment {Section \ref{banalysis}}
                    \State {Apply morphological operations}
                    \State {Merge small blobs}
                    \State {Delete unlikely regions}
        \EndFor
                    
                    \For{Each KCF tracker} 
                    \State {Find best matching blob (section \ref{ostatus})}
                    \EndFor
            \State {Track objects based on current state}
            \If {State == Tracked}
            \State {Adapt scale and update KCF tracker (section \ref{strack})} 
            \ElsIf {State == Occluded}
            \State {Track in straight-through manner (section \ref{soccluded})} 
            \ElsIf {State == New object}
            \State {Create a new KCF tracker (section \ref{snew})} 
            \ElsIf {State == Invisible or exited object}
            \State {Delete or save trajectory (section \ref{sexited})} 
            \EndIf
    \EndFor
    \EndProcedure
  \end{algorithmic}
  \label{algo1}
\end{algorithm}

\section{Experiments}
\label{experiments}
\begin{figure}
\centering
\subfigure[]{\includegraphics[width=4.0cm]{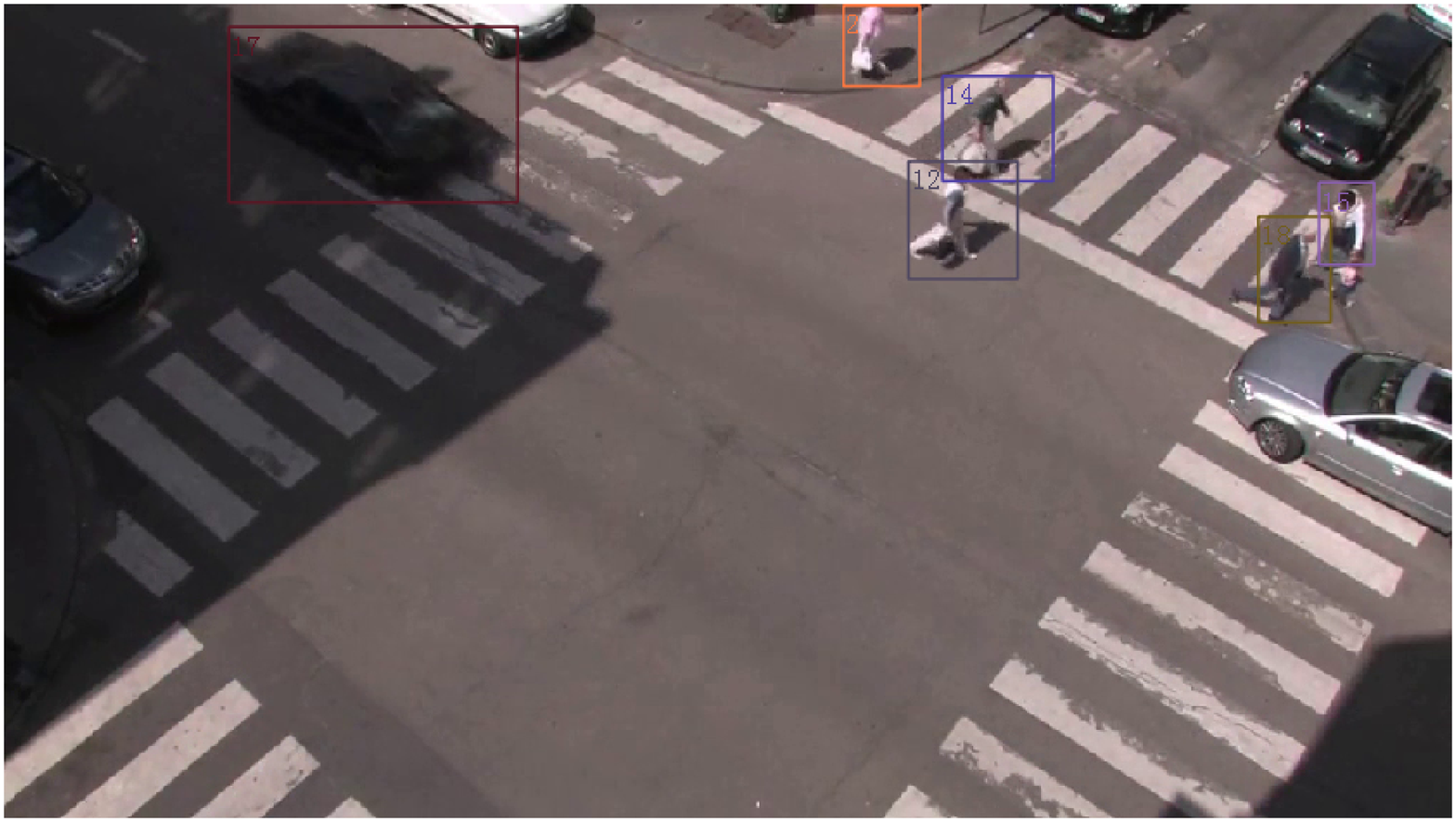}}
\subfigure[]{\includegraphics[width=4.0cm]{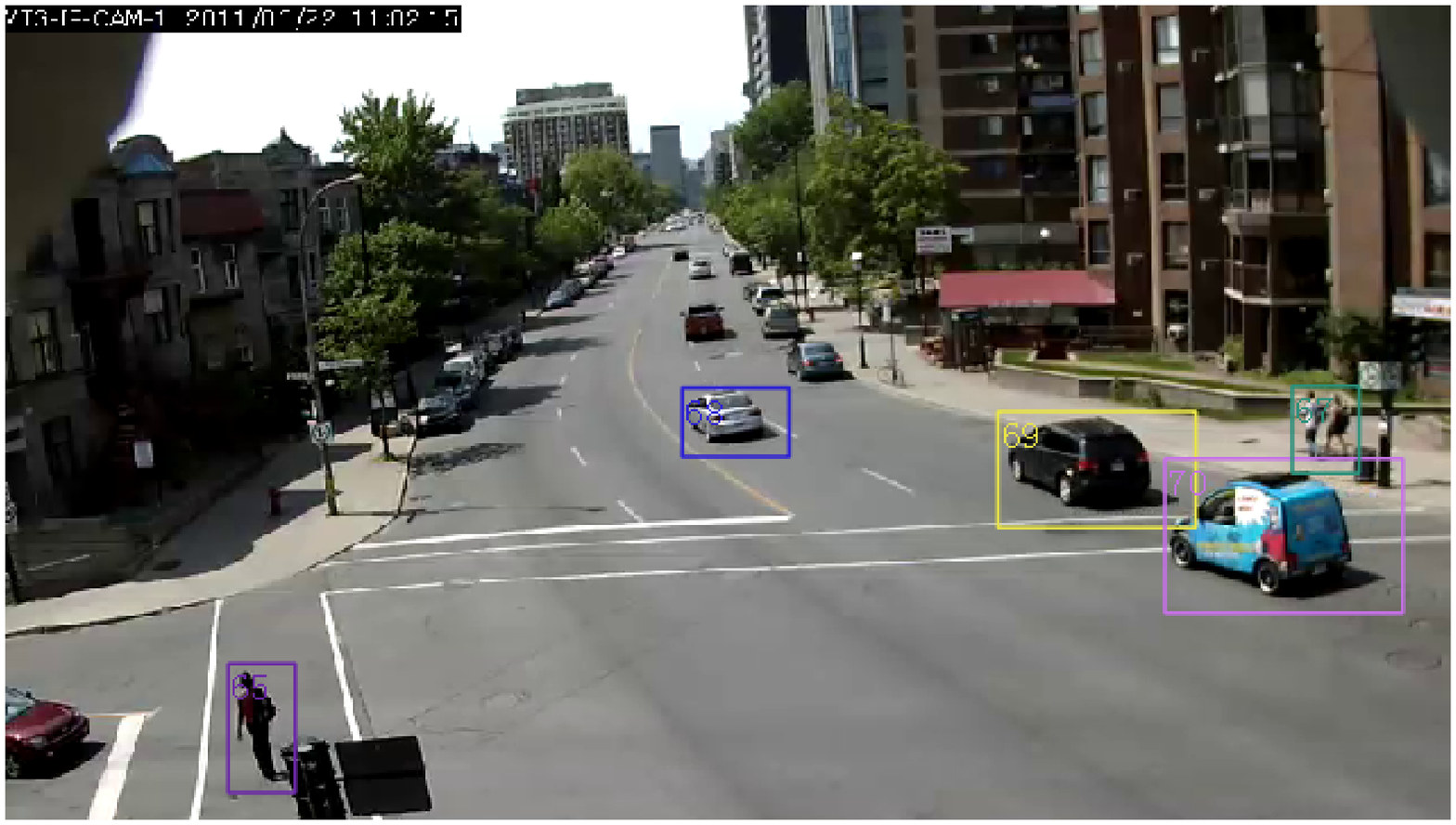}}
\subfigure[]{\includegraphics[width=4.0cm]{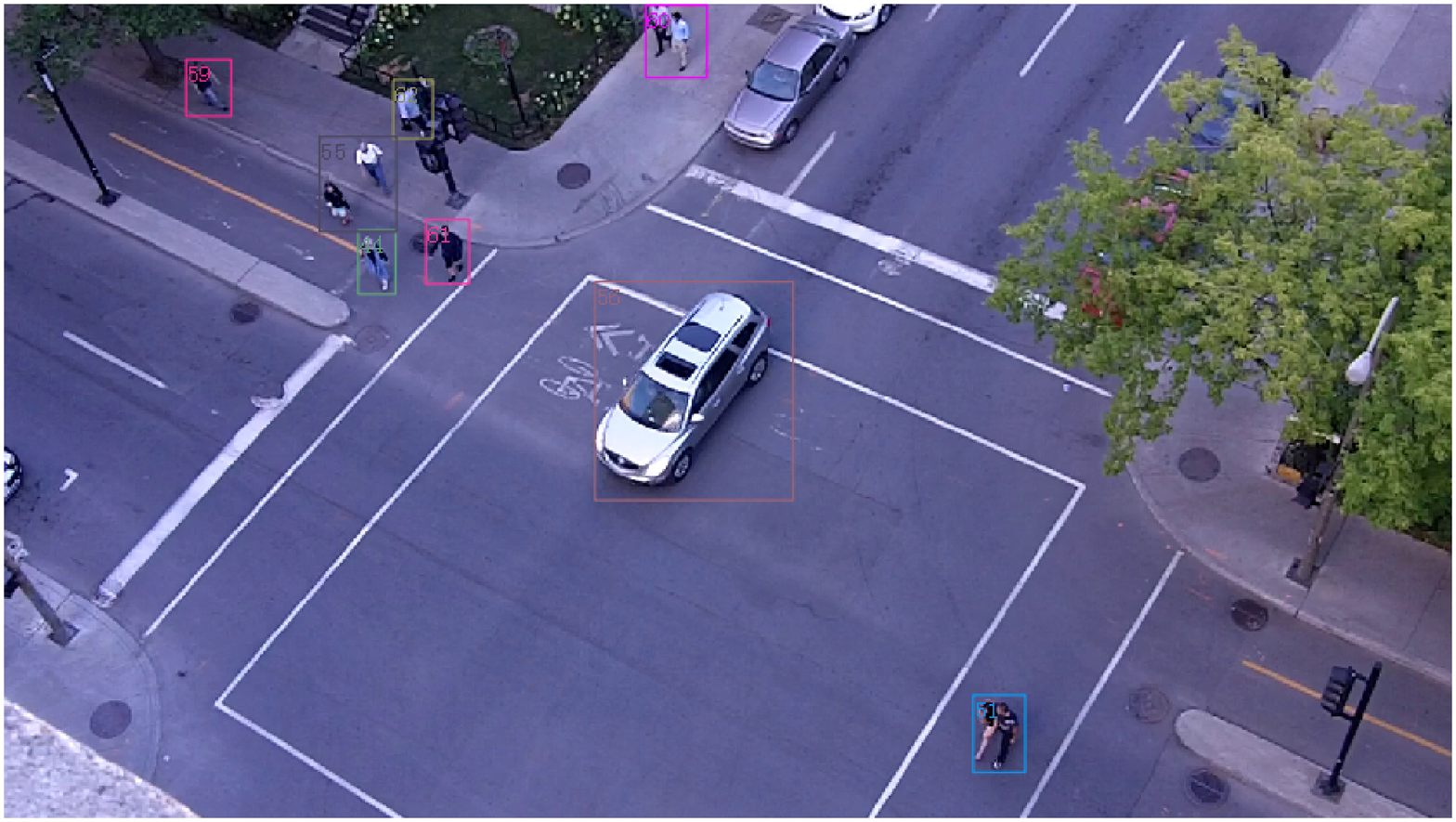}}
\subfigure[]{\includegraphics[width=4.0cm]{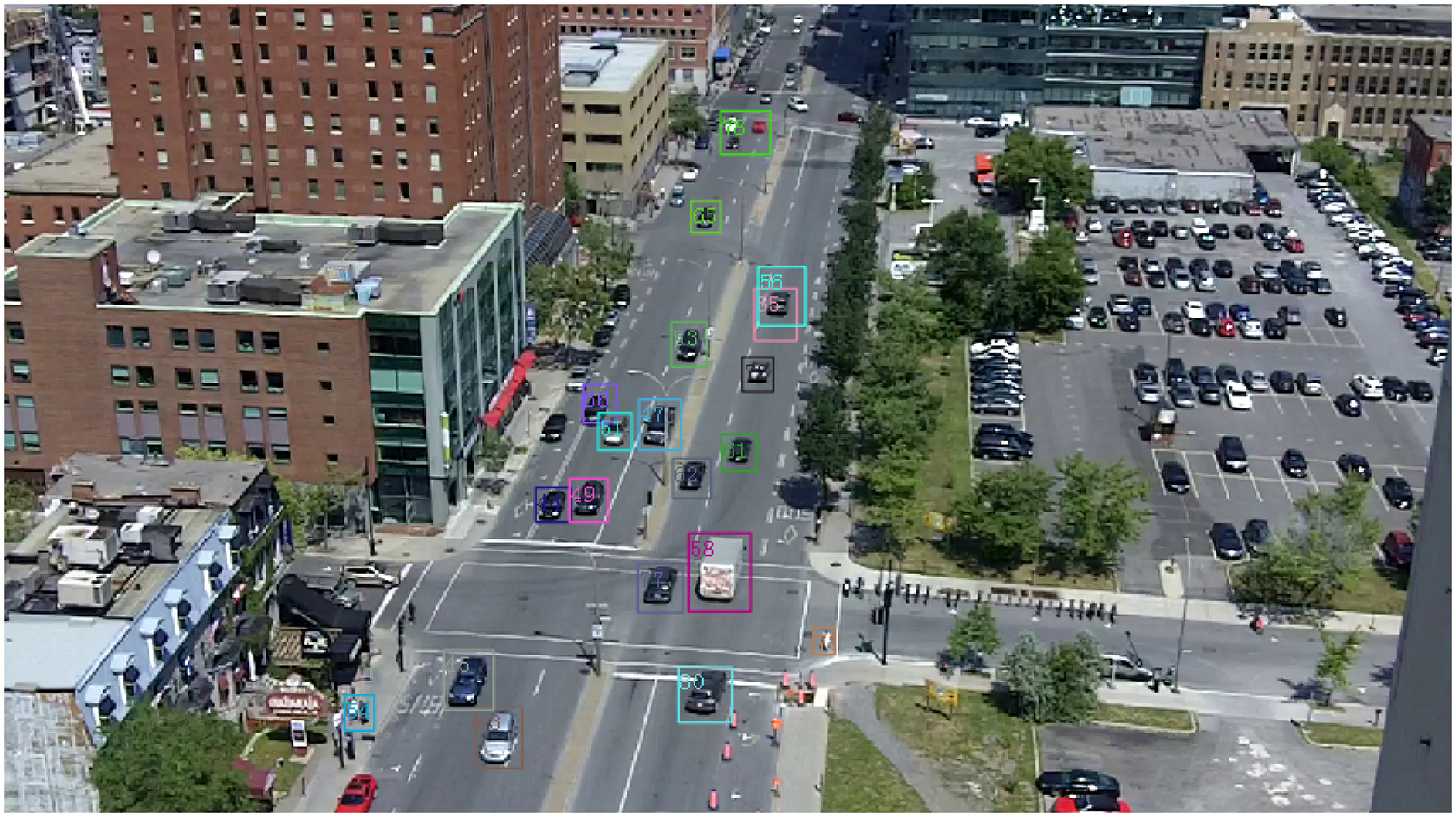}}
\caption{Samples results of our method on the four videos of the Urban Tracker dataset \cite{JodoinITS2016}. (a) Rouen, (b) Sherbrooke, (c) St-Marc, (d) Rene-Levesque} 
\label{fig1}
\end{figure}

To verify the efficiency of our proposed method, we tested our algorithm on four challenging video sequences from a publicly available dataset \cite{JodoinITS2016}. Sample frames of these four videos are shown in figure \ref{fig1}. These videos include cars, cyclists, pedestrians, trucks and buses. The dataset provides tracks that are annotated for all objects that are sufficiently large. We evaluate our method (named MKCF, for multiple KCF) by comparing it with three other trackers: Urban Tracker (UT) \cite{JodoinITS2016}, Traffic Intelligence (TI) \cite{saunier_feature_based_2006} and the tracker of \cite{mendes15vehicle} (Mendes et al.) that we implemented based on the information provided in their paper. Result shows that our algorithm is competitive and promising.

Our method was implemented in C++ and can be downloaded from \url{https://github.com/iyybpatrick/MKCF}.

\subsection{Parameters and performance metrics}
To test our method, we varied two parameters for each video. Blob size $T_r$ is a threshold for filtering regions of small size and the centroid distance $T_c$ is for merging segmented regions in the foreground (see section \ref{banalysis}). The thresholds we adopt for each video are listed in table \ref{table2}. $T_{ol}$ and $T_{oh}$ are fixed to 1.4 and 1.8, respectively. The other tested methods also used different parameter sets for each video.

\begin{table}
\centering
\begin{tabular}{|c|c|c|}
\hline
\multirow{2}{*}{Video}&\multicolumn{2}{c|}{Parameters}\\
\cline{2-3}
& $T_r$ & $T_c$\\
\hline\hline
Sherbrooke&23&44\\
\hline
Rouen&41&63\\
\hline
St-Marc&35&55\\
\hline
Rene-Levesque&20&24\\
\hline
\end{tabular}
\caption{Parameters for each video}
\label{table2}
\end{table}

\begin{table*}[tbp]
\centering
\begin{tabular}{|c|c|c|c|c|c|c|c|c|c|}
\hline
 & & \multicolumn{2}{c|}{MKCF (Ours)} &\multicolumn{2}{c|}{UT} &\multicolumn{2}{c|}{TI}&\multicolumn{2}{c|}{Mendes et al.}\\
\hline
Video &Type &MOTA&MOTP&MOTA&MOTP&MOTA&MOTP&MOTA&MOTP\\
\hline
\multirow{3}{1.2cm}{Sherbrooke}
                & Cars & 0.789 &11.82 & \textbf{0.887} &\textit{10.59} &\textit{0.825} &\textbf{7.42}&0.707  &13.21\\
                & Pedestrians &\textit{0.671} &\textit{7.63} & \textbf{0.705} &\textbf{6.61} &0.014 &11.98 &0.601  &7.97\\
                 &All Objects& \textit{0.763} &10.55 & \textbf{0.787}&\textit{8.64} &0.3841 &\textbf{7.54}&0.695&9.95\\
\hline
\multirow{4}{1cm}{Rouen}
                & Cars & 0.813   & 14.86  & \textit{0.896}    &\textbf{9.73}   & 0.185  & 66.69 &\textbf{0.918} & \textit{11.41}\\
                & Pedestrians & \textit{0.804}   & \textit{14.81}  & \textbf{0.830}   &\textbf{13.77}   & 0.647  &  20.04 &0.672 &14.88 \\
                & Cyclists &  \textit{0.890}  &14.67   & \textbf{0.927}   & 14.13  & 0.869  & \textbf{13.11} &0.881 &\textit{13.67} \\
                & All Objects &  \textit{0.813}  & 14.86  & \textbf{0.844}   & \textbf{13.19}   & 0.589  &24.20 &0.718 &\textit{14.29} \\
\hline
\multirow{3}{1cm}{St-Marc}
                & Cars & 0.590   & 12.54  &  \textbf{0.889}  & \textbf{10.90}  & -0.178  & 38.99 &\textit{0.713}&\textit{11.44}\\
                & Pedestrians & \textbf{0.834}   & 7.35  &  \textit{0.730}   &  \textbf{5.05} & 0.693  & 10.44 &0.505 &\textit{7.09} \\
                & Cyclists & \textit{0.975}   & \textit{6.33}  & \textbf{0.989}   & 6.39  & 0.895  & 7.46 &0.935 &\textbf{6.30} \\
                & All Objects &  \textbf{0.825}  & 7.93  & \textit{0.764}   & \textbf{5.99} & 0.602  & 14.58 &0.560 &\textit{7.71} \\
\hline
\multirow{3}{1cm}{Rene-Levesque}
                & Cars &  \textbf{0.855}  & \textbf{2.72}  & \textit{0.796}   & \textit{3.04}  &  0.547  & 5.23  &0.163  &7.09 \\
                & Cyclists &  \textbf{0.291}  & 2.77  &  \textit{0.232}  &  \textbf{2.20} & \textit{0.232}  & 3.14  &0.216 &\textit{2.26} \\
                & All Objects & \textit{0.572}   & \textbf{2.74}  & \textbf{0.723}   & 2.98  &  0.503  & 5.10 &0.402&\textit{2.95} \\
\hline
\end{tabular}
\caption{Results on the Urban Tracker dataset. MOTP values are in pixels. MOTA should be high, while MOTP should be low for the best performances. \textbf{Boldface}: best result, \textit{Italic}: second best.}
\label{tresults}
\end{table*}

As for the evaluation part, we use the tools provided with the Urban Tracker dataset to calculate CLEAR MOT metrics \cite{clear_mot_2008}.  MOTA is for calculating the multiple object tracking accuracy, which is evaluated based on miss rate, false positive and ID changes. MOTP is for calculating the multiple object tracking precision. It measures the average precision of instantaneous object matches. UT, MKCF and Mendes et al. methods all used the same background subtraction results provided the dataset.

\subsection{Results and analysis}
Quantitative results on the Urban Tracker dataset are provided in table \ref{tresults}. For the Sherbrooke video, our method is competitive with UT and is the second best for the MOTA. In this video, there is a traffic light at the center of the frame. Cars and pedestrians from different directions stop alternatively. Our method often considers two or more objects as one because objects are merged when they first appear in the video. This problem is hard to address because background subtraction fails to segment occluding objects. Moreover, again because of background subtraction, redundant KCF trackers may be created on segmentation of one car because when a car starts moving after a long stop, sometimes it is fragmented into many blobs (see Figure \ref{fig4}a and b). It is difficult to distinguish redundant KCF trackers from many KCF trackers tracking objects in occlusion (see Figure \ref{fig4}c and d). Therefore, redundant trackers cannot be discarded because we are not sure whether they are segments of one object or several isolated objects moving in the same direction. A more complex scheme would be required to handle this case. 

For the Rouen video, our method also ranks second behind UT. For the St-Marc video, although the score of cars is much lower than that of other videos, our proposed method outperforms UT globally and in tracking the pedestrians. Results are lower for cars because more than one KCF trackers are assigned to several segments of one object when the object starts moving, which is the same problem discussed for the Sherbrooke video. This problem affects mostly cars.

For the Rene-Levesque video, our method exhibits the best MOTA for cars and cyclists. However, considering all the objects in the scene, our method performs second because pedestrians in this video are very small. Setting a smaller $T_r$ would improve the performance for tracking pedestrians, but the overall performance will decrease because many environment disturbance cannot be eliminated.

Globally, our method is competitive with the state-of-the-art tracker UT despite not using a complex data association scheme. We just select between background subtraction blobs and KCF bounding boxes. This shows that a state-of-the-art tracker like KCF can help multiple object tracking as demonstrated with performances close to UT. With a more complex data association scheme, the fragmentation problem limitations when initializing a new KCF tracker could be addressed. This problem occurs essentially when an object stops for a long time in the scene. It does not occur for active KCF trackers. Thus, for objects that do not stop for a long time, the use of the KCF tracker is beneficial for handling temporary fragmentation.  

Finally, we also evaluated the speed of our proposed method. Computation times are provided in table \ref{table3}. Reported processing times were obtained on a computer with an intel Core i5-6267U running at 2.9Ghz with 8 Gigabytes of RAM. 

\begin{table*}
\centering
\begin{tabular}{|c|c|c|c|c|}
\hline
Video&Resolution&Number of frames&Processing time (s)&FPS\\
\hline\hline
Sherbrooke&800x600&1001&27.7&36.2\\
\hline
Rouen&1024x576&600&53.1&11.3\\
\hline
St-Marc&1280x720&1000&98.2&10.2\\
\hline
Rene-Levesque&1280x720&1000&53.8&18.6\\
\hline
\end{tabular}
\caption{Processing times of MKCF. FPS: Frames per second}
\label{table3}
\end{table*}

\section{Conclusion}
\label{conclusion}
In this paper, we presented a multiple object tracker that combines a visual object tracker with background subtraction. Background subtraction is applied to extract moving objects and get their scale and size in combination with KCF outputs, while KCF is used for data association and to handle fragmentation and occlusion problems. As a result, KCF and background subtraction help each other to take tracking decision at every frame. This mixed strategy enables us to get competitive results and reduce errors under complex environment. The advantage of our method is that it addresses the occlusion problem by tracking object individually inside blobs of merged objects. Backtracking is not necessary, neither is the use of an explicit region-based segmentation. 

Future work is to design methods to handle stopping objects. Because of background subtraction, they are progressively integrated into the background model. So, when they start moving again they are considered as one or many new objects. Furthermore, we should better distinguish exiting objects from stopping objects.
\bibliographystyle{IEEEtran}
\bibliography{KCFTracker.bib}

\begin{thebibliography}{10}\itemsep=-1pt

\bibitem{Aslani2013}
S.~Aslani and H.~Mahdavi-Nasab.
\newblock Optical flow based moving object detection and tracking for traffic
  surveillance.
\newblock {\em International Journal of Electrical, Robotics, Electronics and
  Communications Engineering}, 7(9):773--777, 2013.

\bibitem{clear_mot_2008}
K.~Bernardin and R.~Stiefelhagen.
\newblock Evaluating multiple object tracking performance: the {CLEAR} {MOT}
  metrics.
\newblock {\em J. Image Video Process.}, 2008:1:1–1:10, jan 2008.

\bibitem{Beymer1997}
D.~Beymer, P.~McLauchlan, B.~Coifman, and J.~Malik.
\newblock A real-time computer vision system for measuring traffic parameters.
\newblock In {\em Computer Vision and Pattern Recognition, 1997. Proceedings.,
  1997 IEEE Computer Society Conference on}, pages 495--501, 1997.

\bibitem{coifman_real-time_1998}
B.~Coifman, D.~Beymer, P.~McLauchlan, and J.~Malik.
\newblock A real-time computer vision system for vehicle tracking and traffic
  surveillance.
\newblock {\em Transportation Research Part C: Emerging Technologies},
  6:271--288, 1998.

\bibitem{ExtendedKCF}
M.~Danelljan, F.~S. Khan, M.~Felsberg, and J.~v.~d. Weijer.
\newblock Adaptive color attributes for real-time visual tracking.
\newblock In {\em Proceedings of the 2014 IEEE Conference on Computer Vision
  and Pattern Recognition}, CVPR '14, pages 1090--1097, Washington, DC, USA,
  2014. IEEE Computer Society.

\bibitem{Fuentes2006}
L.~M. Fuentes and S.~A. Velastin.
\newblock People tracking in surveillance applications.
\newblock {\em Image and Vision Computing}, 24(11):1165--1171, 2006.

\bibitem{HenriquesKCF}
J.~F. Henriques, R.~Caseiro, P.~Martins, and J.~Batista.
\newblock High-speed tracking with kernelized correlation filters.
\newblock {\em IEEE Transactions on Pattern Analysis and Machine Intelligence},
  37(3):583--596, March 2015.

\bibitem{JodoinWACV14}
J.~Jodoin, G.~Bilodeau, and N.~Saunier.
\newblock Urban tracker: Multiple object tracking in urban mixed traffic.
\newblock In {\em {IEEE} Winter Conference on Applications of Computer Vision,
  Steamboat Springs, CO, USA, March 24-26, 2014}, pages 885--892, 2014.

\bibitem{JodoinITS2016}
J.~P. Jodoin, G.~A. Bilodeau, and N.~Saunier.
\newblock Tracking all road users at multimodal urban traffic intersections.
\newblock {\em IEEE Transactions on Intelligent Transportation Systems},
  PP(99):1--11, 2016.

\bibitem{jun_tracking_2008}
G.~Jun, J.~K. Aggarwal, and M.~Gokmen.
\newblock Tracking and segmentation of highway vehicles in cluttered and
  crowded scenes.
\newblock In {\em Proceedings of the 2008 {IEEE} Workshop on Applications of
  Computer Vision}, {WACV} '08, page 1–6, Washington, {DC}, {USA}, 2008.
  {IEEE} Computer Society.

\bibitem{kim_real_2008}
Z.~Kim.
\newblock Real time object tracking based on dynamic feature grouping with
  background subtraction.
\newblock In {\em {IEEE} Conference on Computer Vision and Pattern Recognition,
  2008. {CVPR} 2008}, pages 1--8, 2008.

\bibitem{Lessard_2016_CVPR_Workshops}
A.~Lessard, F.~Belisle, G.-A. Bilodeau, and N.~Saunier.
\newblock The countingapp, or how to count vehicles in 500 hours of video.
\newblock In {\em The IEEE Conference on Computer Vision and Pattern
  Recognition (CVPR) Workshops}, June 2016.

\bibitem{mendes15vehicle}
J.~C. Mendes, A.~G.~C. Bianchi, and A.~R.~P. J{\'u}nior.
\newblock Vehicle tracking and origin-destination counting system for urban
  environment.
\newblock In {\em Proceedings of the International Conference on Computer
  Vision Theory and Applications (VISAPP 2015)}, 2015.

\bibitem{saunier_feature_based_2006}
N.~Saunier and T.~Sayed.
\newblock A feature-based tracking algorithm for vehicles in intersections.
\newblock In {\em The 3rd Canadian Conference on Computer and Robot Vision,
  2006}, pages 59--59, 2006.

\bibitem{KLT}
J.~Shi and C.~Tomasi.
\newblock Good features to track.
\newblock In {\em Computer Vision and Pattern Recognition, 1994. Proceedings
  CVPR '94., 1994 IEEE Computer Society Conference on}, pages 593--600, 1994.

\bibitem{CharlesB14}
P.~St{-}Charles and G.~Bilodeau.
\newblock Improving background subtraction using local binary similarity
  patterns.
\newblock In {\em {IEEE} Winter Conference on Applications of Computer Vision,
  Steamboat Springs, CO, USA, March 24-26, 2014}, pages 509--515, 2014.

\bibitem{Torabi2009}
A.~Torabi and G.~A. Bilodeau.
\newblock A multiple hypothesis tracking method with fragmentation handling.
\newblock In {\em Computer and Robot Vision, 2009. CRV '09. Canadian Conference
  on}, pages 8--15, 2009.

\end{thebibliography}

\end{document}